\documentclass[runningheads]{llncs}
\usepackage{graphicx}
\usepackage{amsmath}
\usepackage{subfigure}
\usepackage{multirow}
\usepackage{wrapfig}
\usepackage{amssymb}
\usepackage[misc,geometry]{ifsym}
\begin{document}
\title{Sample hardness based gradient loss for long-tailed cervical cell detection}
\titlerunning{Sample hardness based gradient loss for long-tailed cervical cell detection}

\author{Minmin Liu\inst{1,2,3} \and   
Xuechen Li \inst{1,2,3,4} \and          
Xiangbo Gao \inst{5}  \and            
Junliang Chen \inst{1,2,3} \and       
Linlin Shen \inst{1,2,3}$^\textrm{(\Letter)}$  \and        
Huisi Wu \inst{1,2,3}}                

\authorrunning{M. Liu et al.}

\institute{Computer Vision Institute, School of Computer Science and Software Engineering, Shenzhen University, China, \and Shenzhen Institute of Artificial Intelligence of Robotics of Society, Shenzhen, China, \and Guangdong Key Laboratory of Intelligent Information Processing, Shenzhen University, Shenzhen 518060, China, \\\email{llshen@szu.edu.cn}\\ \and National Engineering Laboratory for Big Data System Computing Technology, Shenzhen University, Shenzhen 518060, PR China \and University of California, Irvine\\}

\maketitle              
\begin{abstract}
Due to the difficulty of cancer samples collection and annotation, cervical cancer datasets usually exhibit a long-tailed data distribution. When training a detector to detect the cancer cells in a WSI (Whole Slice Image) image captured from the TCT (Thinprep Cytology Test) specimen, head categories (e.g. normal cells and inflammatory cells) typically have a much larger number of samples than tail categories (e.g. cancer cells). Most existing state-of-the-art long-tailed learning methods in object detection focus on category distribution statistics to solve the problem in the long-tailed scenario, without considering the ``hardness'' of each sample. To address this problem, in this work we propose a Grad-Libra Loss that leverages the gradients to dynamically calibrate the degree of hardness of each sample for different categories, and re-balance the gradients of positive and negative samples. Our loss can thus help the detector to put more emphasis on those hard samples in both head and tail categories. Extensive experiments on a long-tailed TCT WSI image dataset show that the mainstream detectors, e.g. RepPoints, FCOS, ATSS, YOLOF, etc. trained using our proposed Gradient-Libra Loss, achieved much higher (7.8\%) mAP than that trained using cross-entropy classification loss.

\keywords{Long-tailed learning  \and Object detection \and Cervical cancer.}
\end{abstract}

\section{Introduction}
Cervical cancer is the fourth most frequently diagnosed cancer and the fourth leading cause of cancer death in women~\cite{sung2021global}. Early diagnosis and screening of cervical cancer can effectively help its treatment. To solve the error-prone, tedious, and time-consuming problems of manual analysis of cervical smears, deep learning based CAD (Computer-Aided Diagnosis) has been introduced to cervical cancer screening. However, due to the difficulty of cancer samples collection and the cost of annotation, the number of cancer samples is far less than that of normal samples, which shows a typical long-tailed distribution and leads to a long-tailed class imbalance problem.

In this paper, we are mainly using object detectors to detect cancer cells in a WSI (Whole Slide Image) image captured from a TCT (Thinprep Cytology Test) specimen. The problem of training cell detectors on a long-tailed dataset mainly comes from two aspects. First, the categories are extremely imbalanced, which will cause the loss contributions of the tail classes to be easily overwhelmed by the head classes. Second, for the object detection framework, the background forms a large number of easy negative samples, which will also overwhelm the training process and degrade the training performance. Note that for a particular category, the samples belonging to the category are positive samples, while the samples of all the other categories and the background are negative samples. Most of the existing methods addressing long-tailed problem require additional statistics support~\cite{cui2019class,gupta2019lvis,han2005borderline,he2009learning,huang2016learning,li2020overcoming} (e.g. data distribution statistics), or tedious operations~\cite{kang2019decoupling,wang2020devil,li2020overcoming} (e.g. fine-tuning and handcrafted head-tail class division). Data re-sampling methods~\cite{gupta2019lvis,han2005borderline,he2009learning} require the acquisition of pre-computed data distribution statistics, which may have the risks of over-fitting for tail classes and under-fitting for head classes. Loss re-weighting methods~\cite{cui2019class,huang2016learning} also require data distribution statistics for up-weighting the tail classes and down-weighting the head classes at class level. Decoupled training schema~\cite{kang2019decoupling,wang2020devil} decouples representation and classifier learning but requires an extra fine-tuning stage. BAGS~\cite{li2020overcoming} divided all categories into several groups according to data distribution statistics during the training stage but the handcrafted division may block the sharing of information between the head and tail classes. In general, most existing methods focus on data distribution statistics to solve the class imbalance problem in the long-tailed scenario at class level, without taking into account the ``\textbf{hardness}'' of each sample at sample level. In fact, we cannot ignore the diversity of samples. A sample has different hardness (easy or hard) for the classification of different categories. It can be a hard positive sample of its category and a hard negative sample of other categories at the same time. Thus, there might be easy samples belonged to the tail classes get incorrectly up-weighted, or the opposite. To address this problem, we propose a Grad-Libra Loss that leverages the gradients to dynamically calibrate the degree of hardness of each sample for different categories, and re-balance the gradients of positive and negative samples.

To sum up, our contributions are as follows: (1) We propose a new perspective to describe the long-tailed samples, which defines the concept of sample hardness and calibrates the degree of hardness by gradients. (2) We provides a unified framework to take the hardness into account for both head and tail classes and present a novel Gradient Libra Loss that employs the gradients to adaptively re-weight samples of different hardness. (3) We conduct comprehensive experiments using a long-tailed cervical cell image dataset. Our method consistently surpasses most existing methods and obtains a 7.8\% mAP gain over the baseline model, increasing the AP of frequent, common, and rare categories by 7.3\%, 5.6\%, and 8.4\%, respectively. 

\begin{figure}[t]
\includegraphics[width=\textwidth]{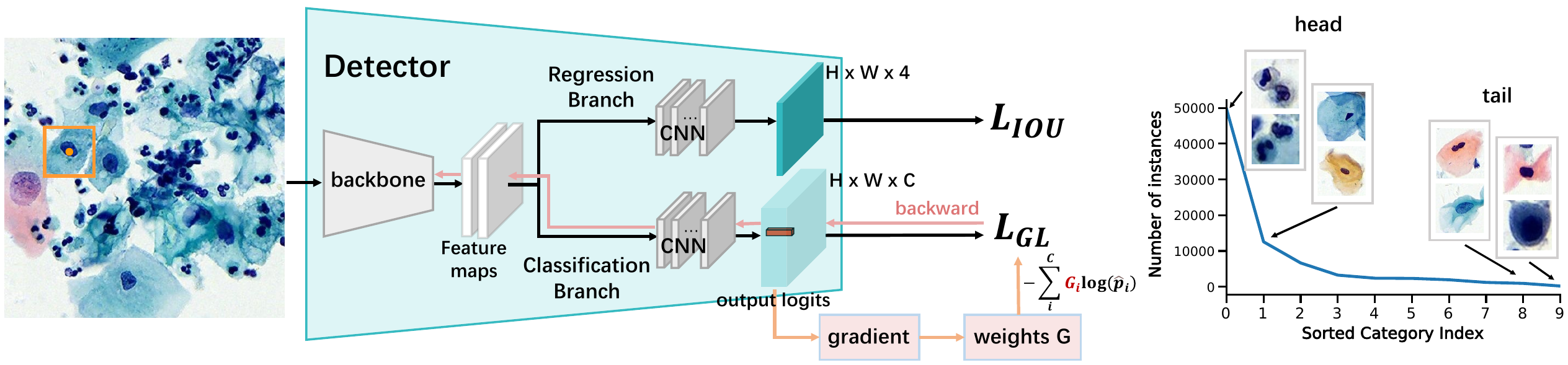}
\caption{The framework of the cell detector. We propose a novel Grad-Libra Loss for the classification branch. In the right part, we visualize the data distribution of the long-tailed dataset along with example instances.} \label{fig1}
\end{figure}

\section{System Framework}
Fig.\ref{fig1} shows the framework of our cell detector, which is based on mainstream object detectors like RepPoints, FCOS, ATSS, YOLOF, etc. During training, a batch of TCT WSI image patches is sampled from a dataset with the long-tailed distribution of cell categories and fed to the detector for loss calculation and gradient backpropagation. While Smooth L1 Loss~\cite{chen2021you,tian2019fcos,yang2019reppoints,zhang2020bridging} is usually used to train the regression of the bounding box, the cross-entropy loss is the main choice for the training of the classification branch. For the classification branch, we employ multiple binary classifiers for multi-class classification and design a novel Gradient Libra Loss. It employs the gradient to calibrate the degree of hardness of each sample at sample level and takes hardness as the weight term of the original cross-entropy loss function.

\section{Gradient Libra Loss}

\begin{figure}[t]
\includegraphics[width=\textwidth]{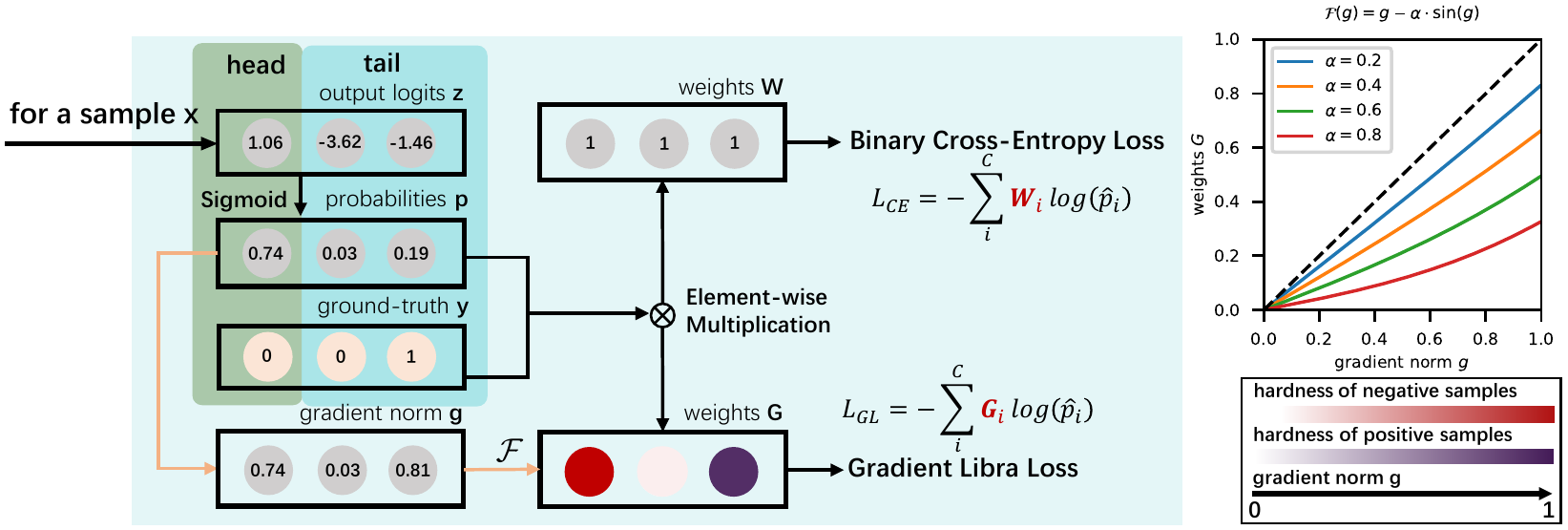}
\caption{Visualization of the detailed procedure of the classification loss function-Gradient Libra Loss, which takes hardness as the weight term of the original cross-entropy loss function. Note that the darker the color in the gradient bar, the stronger the hardness.} \label{fig2}
\end{figure}

To better explain our loss, we firstly analyze the connection between the gradients and positive-negative imbalance. Suppose we have a batch of samples $\mathcal{I}$ and their features for the classification branch, and their output logits are used to represent the attributes (e.g. easy or hard) of the samples. As shown in Fig.\ref{fig2}, for a sample $\mathbf{x}$ of class $k$, $\mathbf{z}=$ $[z_{1}, z_{2},...,z_{C}]^{T}$ and $\mathbf{p}=[p_{1}, p_{2},...,p_{C}]^{T}$ are the output logits and probabilities for each class, respectively. The ground-truth label $\mathbf{y}$ of the sample is an one-hot vector, in which $y_{i}\in \{0,1\}, 1 \leq i \leq C$ and $y_{k}=1$ and $y_{i}=0$ $(i \neq k)$. Its gradient with regard to the output logits $\mathbf{z}$ in the original cross-entropy $L_{CE}$ is as follows:
\begin{equation}
\frac{\partial L_{CE}}{\partial z_{i}} = \begin{cases}
\ p_i - 1, & if \ i=k \\
\ p_i, & if \ i \neq k \\
\end{cases}
\label{eq1}
\end{equation}

Eq.\ref{eq1} means that for the sample $\mathbf{x}$ of class $k$, it obtains the encouraging gradient $p_{i}-1$ as positive sample but gets the penalty gradient $p_i$ as negative sample for other class $i(i \neq k)$. As presented in~\cite{tan2021equalization}, we choose the ratio $r$ of cumulative gradients between positive and negative samples as an indicator to measure whether each category classifier is in a positive-negative balanced training state. For iteration $t + 1$, the ratio $r$ is defined as $r=\sum_{t^{*}=0}^{t}|\nabla_{z_i}^{+}(L)| \backslash \sum_{t^{*}=0}^{t}|\nabla_{z_i}^{-}(L)|$. The gradients of positive samples $\nabla_{z_i}^{+}(L)$ and gradients of negative samples $\nabla_{z_i}^{-}(L)$ of output logit $z_i$ are formulated as:$\nabla_{z_i}^{+}(L) = \frac{1}{|\mathcal{I}|}\sum_{n \in \mathcal{I} } y_i^n(p_i^n-1)$,  $\nabla_{z_i}^{-}(L) = \frac{1}{|\mathcal{I}|}\sum_{n \in \mathcal{I} } (1-y_i^n)p_i^n$. In general, for the head classes, the gradients of positive and negative samples have similar magnitudes, and the ratio $r$ is close to or greater than 1. For the tail classes, the gradients of positive samples are often overwhelmed by the gradients of negative samples, resulting in the ratio $r$ close to 0.


 Our intention is to adopt the gradients to calibrate the hardness of the samples and re-weight the loss function. We define $\mathbf{g}=[g_{1}, g_{2},...,g_{C}]^{T}$, where $g_{i} = |\frac{\partial L_{CE}}{\partial z_{i}}|$. $\mathbf{g}$ is equal to the norm of gradient w.r.t output logits $\mathbf{z}$. $\mathbf{g}$ passes through a monotonically increasing function $\mathcal{F}:\mathbf{g} \to \mathbf{G}$, and $\mathcal{F}$ is defined as:
\begin{equation}
\mathcal{F}(\mathbf{g}) = \mathbf{g} - \alpha \cdot \sin(\mathbf{g}) 
\label{eq3}
\end{equation}

$\mathbf{G}$ acts as the weight term of $L_{CE}$. $\alpha \in (0, 1]$ is a modulating factor to control the importance of samples. The integrated Grad-Libra Loss is expressed as follows:
\begin{equation}
L_{GL}= -\sum_{i}^{C} \mathbf{G}_{i} \log(\hat{p_{i}}), \
\hat{p_{i}} = \begin{cases}
\ p_i, & if \ i=k \\
\ 1-p_i, & if \ i \neq k \\
\end{cases}
\label{eq4}
\end{equation}

Decouple the $L_{CE}$ according to the positive and negative level:
$L_{CE}^{+} = -y_{i}\log(p_{i})$ and $L_{CE}^{-} = -(1-y_{i})\log(1-p_{i})$. $L_{CE}^{+}$ and $L_{CE}^{-}$ represent the positive and negative loss term in class $i$ (omit the class index $i$ for brevity), respectively. We set different modulating factors for positive samples and negative samples, i.e., $\alpha^{+}$ and $\alpha^{-}$. Grad-Libra Loss in class $i$ is decoupled as follows:
\begin{equation}
\begin{cases}
\ L_{GL}^{+} =  G^{+} \cdot L_{CE}^{+}  \\
\ L_{GL}^{-} =  G^{-} \cdot L_{CE}^{-} \\
\end{cases}, \
\begin{cases}
\ G^{+} = (1-p_{i}) - \alpha^{+}sin(1-p_{i})\\
\ G^{-} = p_{i} - \alpha^{-}sin(p_{i})\\
\end{cases}
\label{eq5}
\end{equation}

The unified framework of Grad-Libra Loss is written as:
\begin{equation}
L_{GL} = -\sum_{i}^{C}L_{GL}^{+} + L_{GL}^{-} = -\sum_{i}^{C}{\mathcal{F}(\mathbf{g};\alpha^{+}}) \cdot L_{CE}^{+} + {\mathcal{F}(\mathbf{g};\alpha^{-}}) \cdot L_{CE}^{-}
\label{eq6}
\end{equation}

The gradient norm $\mathbf{g}$ is nonlinearly transformed by the function $\mathcal{F}$ and then used as the weight of the loss function. In our unified framework, an appropriate function $\mathcal{F}$ can achieve excellent performance.

\textbf{Implication 1} (Positive-Negative Balance). As shown in Eq.\ref{eq5}, $\alpha$ adjusts the weights of positive and negative samples. The overall weight $\mathbf{G}$ will increase according to the decrease of the $\alpha$, as shown in the top right of Fig.\ref{fig2}. The overall importance of $G^{+}$ and $G^{-}$ can be balanced by $\alpha^{+}$ and $\alpha^{-}$. Proper adjustment of $\alpha^{+}$ and $\alpha^{-}$ can promote positive-negative balanced training. Notably, different detectors have different degrees of positive-negative imbalance. In order to achieve a more satisfactory effect, it is necessary to re-adjust $\alpha^{+}$ and $\alpha^{-}$ when using different detectors.

\textbf{Implication 2} (Hardness). $\mathbf{G}$ focuses on the hardness (easy or hard) of samples. $G^{+}$ and $G^{-}$ adjust the weights of easy-hard positive and negative samples. When a sample's probability $p_i$ for a category is closer to the ground-truth, it is more likely to be an easy sample of that category. The gradient norm $g_{i}$ becomes smaller, and leads to a smaller weight $G_{i}$. Therefore, we use $\mathbf{G}$ to represent the hardness of the samples. As shown in Fig.\ref{fig2}, we visualize an example of loss computing for a tail sample and employ different colors to indicate the hardness. For a sample $\mathbf{x}$ belonging to a certain tail class $i$, its output logits $\mathbf{z}$ is transformed by the sigmoid function into estimated probabilities $\mathbf{p}$. The original cross-entropy loss does not consider the hardness of samples for different categories (the weight vector is set to 1), which seriously reduces the discrimination ability to of easy or hard samples. For the proposed Gradient Libra Loss, it calculates the gradient norm of the sample, obtains the degree of hardness through the function of $\mathcal{F}$, and uses it as $\mathbf{G}$ to weight the original cross-entropy loss. Through the weight $\mathbf{G}$ we can estimate the hardness of the samples for different classes. For example, as the positive sample (ground-truth is 1) of a certain tail class $i$, its probability of 0.19 is very small, so it tends to be a hard-positive sample of its own category (the weight $G_i$ is marked by dark purple). As the negative sample (ground-truth is 0) of other categories, it has different probabilities and therefore different degrees of hardness (darker color represents higher hardness).

\section{Experiments}
\subsection{Dataset and Evaluation Metrics}
Public cervical cancer datasets~\cite{plissiti2018sipakmed,jantzen2005pap} are small in scale, balanced in categories, which cannot represent the real distribution in clinical practice. Thus, we establish the CCA-LT dataset \renewcommand{\thefootnote}{1}\footnote{https://github.com/M-LLiu/Grad-Libra} that is closer to the actual long-tailed data distribution. We visualize the data distribution of the CCA-LT dataset along with example instances in Fig.\ref{fig1}. The CCA-LT dataset was captured by a whole-slide scanning machine. The specimens are prepared by Thinprep Cytology Test method with H\&E staining~\cite{zhang2014automation} to give the image resolution of about 80000×60000 pixels. The dataset is divided into training-validation set and test set (7:3). The training-validation set is overlap-cropped for data augmentation, while the test set is not. Then we crop each whole slide digital image into patches of 512×512 pixels for training. To evaluate detection performance, we follow the PASCAL VOC evaluation criteria~\cite{everingham2010pascal}, i.e., mean average precision (mAP). We also report AP for each category, which is calculated with IoU threshold of 0.5. The AP and mean recall for rare, common, and frequent categories are denoted as {AP$_r$}, {AP$_c$}, {AP$_f$}, {mR$_r$}, {mR$_c$}, and {mR$_f$}, respectively.

\subsection{Implementation Details}
For a fair comparison, all experiments are performed on MMDetecion~\cite{chen2019mmdetection} platform in PyTorch~\cite{paszke2019pytorch} framework. We use 4 V100 GPUs for training. We choose the anchor-free detector RepPoints~\cite{yang2019reppoints} with FPN structure~\cite{lin2017feature} as the baseline model. The optimizer is stochastic gradient descent (SGD) with momentum 0.9 and weight decay 0.0001. The initial learning rate is set as 0.002, which is divided by 10 after 8 and 11 epochs. We employ the linear warming up policy~\cite{goyal2017accurate} to start the training and the warm-up ratio is set as 0.001. The model is trained with batch size of 32 for 12 epochs.

\begin{table}[t]
\caption{The performance of Grad-Libra Loss compared with other methods. G means number of groups. * means adding the effect of $\alpha$-balanced factor.}\label{tab2}
\resizebox{\textwidth}{!}{
\begin{tabular}{c|c|cc|ccccccc|c}
\hline
  \multirow{2}{*}{Method} & \multicolumn{1}{c|}{frequent} & \multicolumn{2}{c|}{common} & \multicolumn{7}{c|}{rare}  & \multirow{2}{*}{mAP}\\ \cline{2-11}
 & inflammation & normal & HSIL & atrophy & bare\_nucleus & SCC & trichomonad & LSIL & ASC\_US & ASC\_H & \\ \hline

CBL \cite{cui2019class} & 20.2 & 65.0 & 24.7 & 30.1 & 26.7 & 0.0 & 26.9 & 27.2 & 17.2 & 0.0 & 23.8  \\ 
Seesaw \cite{wang2021seesaw} & 71.3 & 75.4 & 55.9 & 87.2 & 58.7 & 2.1 & 9.1 & \textbf{61.3} & 26.4 & 2.3 & 45.0  \\ 
CE & 72.8 & 76.3 & 57.2 & 88.0 & 56.0 & 6.6 & 16.0 & 59.8 & 18.9 & 1.9 & 45.3 \\ 
EQLv2 \cite{tan2021equalization} & 74.7 & 77.1 & 54.7 & 87.1 & 62.6 & 4.4 & 19.7 & 56.7 & 23.2 & 0.7 & 46.1  \\ 
BAGS (G3) \cite{li2020overcoming} & 80.3 & 77.1 & 61.2 & 87.2 & 69.7 & 0.9 & 22.4 & 55.9 & 21.9 & 0.9 & 47.8  \\ 
BAGS (G2) \cite{li2020overcoming} & \textbf{80.4} & 77.5 & 54.8 & 87.4 & 70.1 & 3.0 & 29.4 & 59.1 & 27.1 & 1.4 & 49.0  \\ 
Focal \cite{lin2017focal} & \textbf{80.4} & 80.8 & 56.2 & 85.0 & 66.8 & 5.6 & 31.6 & 53.9 & 24.0 & 6.0 & 49.0  \\ 
Focal* \cite{lin2017focal} & 79.6 & 81.7 & 60.9 & 87.3 & \textbf{72.7} & 3.5 & 31.2 & 61.1 & 24.5 & 3.4 & 50.6  \\ 
Grad-Libra (ours) & 80.1 & \textbf{83.2} & \textbf{61.6} & \textbf{88.3} & 69.7 & \textbf{10.3} & \textbf{40.6} & 60.2 & \textbf{27.3} & \textbf{9.6} & \textbf{53.1}  \\  \hline
\end{tabular}}
\end{table}
\subsection{Benchmark Results}
We compare the performance of Grad-Libra Loss with other state-of-the-art methods and report the results in Table~\ref{tab2}. Without any bells and whistles, Grad-Libra Loss achieves better results than all other losses and exceeds the CE baseline \textbf{7.8\%} mAP. Without sacrificing the accuracy of the head classes, Grad-Libra Loss brings significant performance gains to the tail classes, e.g. increasing AP$_f$ by 7.3\%, AP$_c$ by 5.6\%, and AP$_r$ by 8.4\%, respectively. We further compare Grad-Libra Loss with recent designs, i.e., Class-balanced Loss (CBL) \cite{cui2019class}, Balanced Group Softmax (BAGS) \cite{li2020overcoming}, Equalization Loss v2 (EQLv2) \cite{tan2021equalization}, and Seesaw Loss (Seesaw) \cite{wang2021seesaw} in Table~\ref{tab2}. Notably, CBL and Seesaw achieve 21.2\% and 0.3\% lower mAP than the CE baseline, respectively. CBL and Seesaw, which require data distribution information at class level or the number of samples in sample level, have poor robustness and drop sharply in performance. We follow BAGS and split the classes into 2 groups (0, 10000), (10000, $+\infty$) and 3 groups (0, 5000), (5000, 10000), (10000, $+\infty$) for group softmax computation, respectively. The division of 2 groups is better, which illustrates the necessity of sharing the head knowledge to reduce confusion between head and tail classes and enhances tail discrimination. Designed to address the extreme easy-hard imbalance problem, Focal Loss~\cite{lin2017focal} can not effectively alleviate the tail classes problem. By employing the $\alpha$-balanced factor in regulating the positive-negative imbalance, Focal* achieves 1.6\% higher mAP, which confirms the importance of focusing on the positive-negative imbalance. In contrast, Grad-Libra Loss employs gradient information to re-balance positive-negative samples of different hardness and achieves the best 53.1\% mAP performance. Grad-Libra Loss exceeds the second-best loss Focal* by 0.5\% AP$_f$, 1.1\% AP$_c$, and 3.2\% AP$_r$ in frequent, common, and rare categories, respectively.

\begin{figure}[t]
    \centering
    \subfigure[]{\includegraphics[scale=0.3]{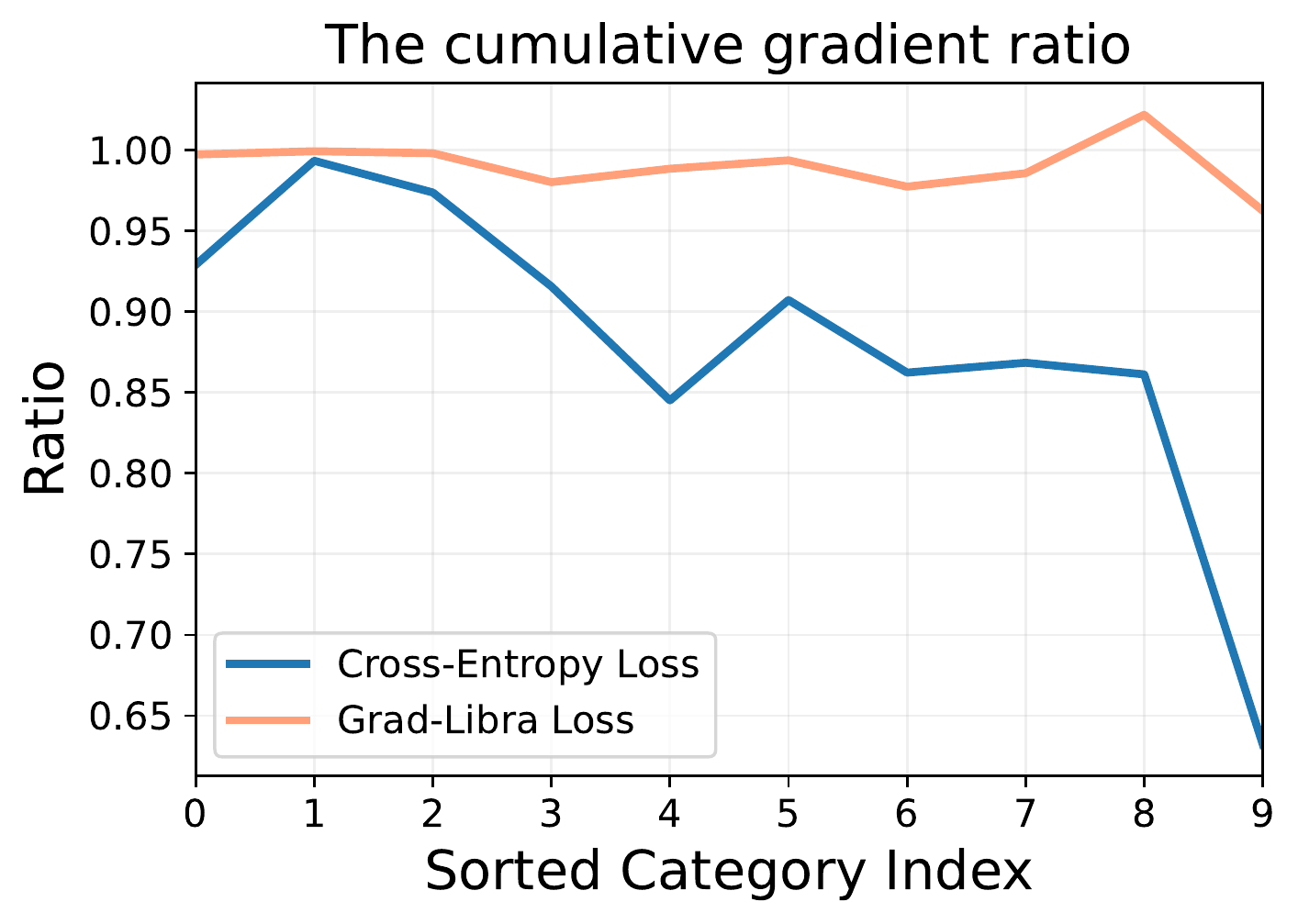}}
    \noindent
    \quad
    \subfigure[]{\includegraphics[scale=0.3]{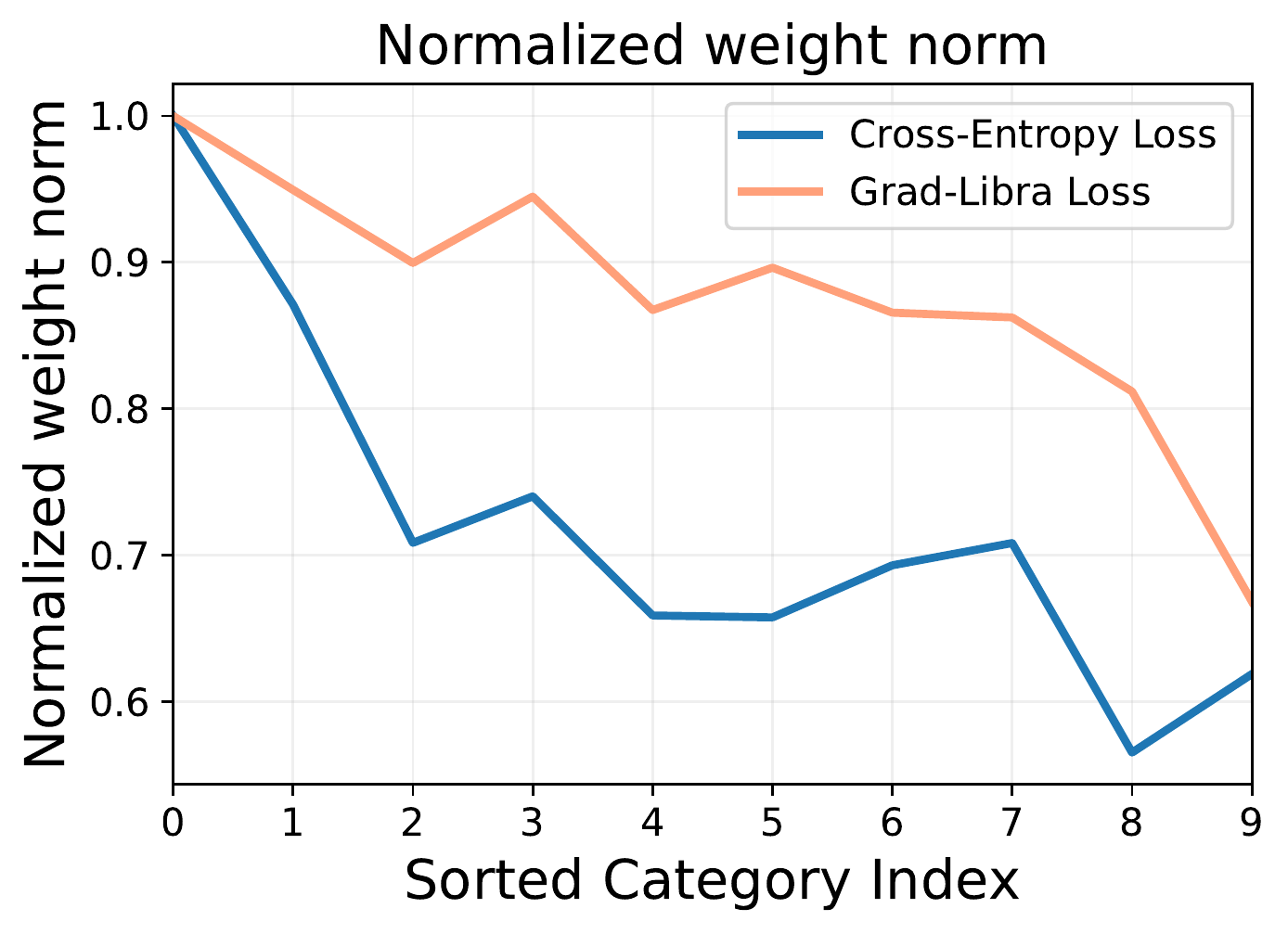}}

    \caption{The comparison of gradient balance between Grad-Libra and Cross-Entropy Loss. (a): The cumulative gradient ratio in the entire training process for each category. (b): The normalized L2 weight norm in the last classifier  layer for each category.}
    \label{fig3}
\end{figure}
\begin{figure}
    \centering
    \subfigure[]{\includegraphics[width=0.35\textwidth]{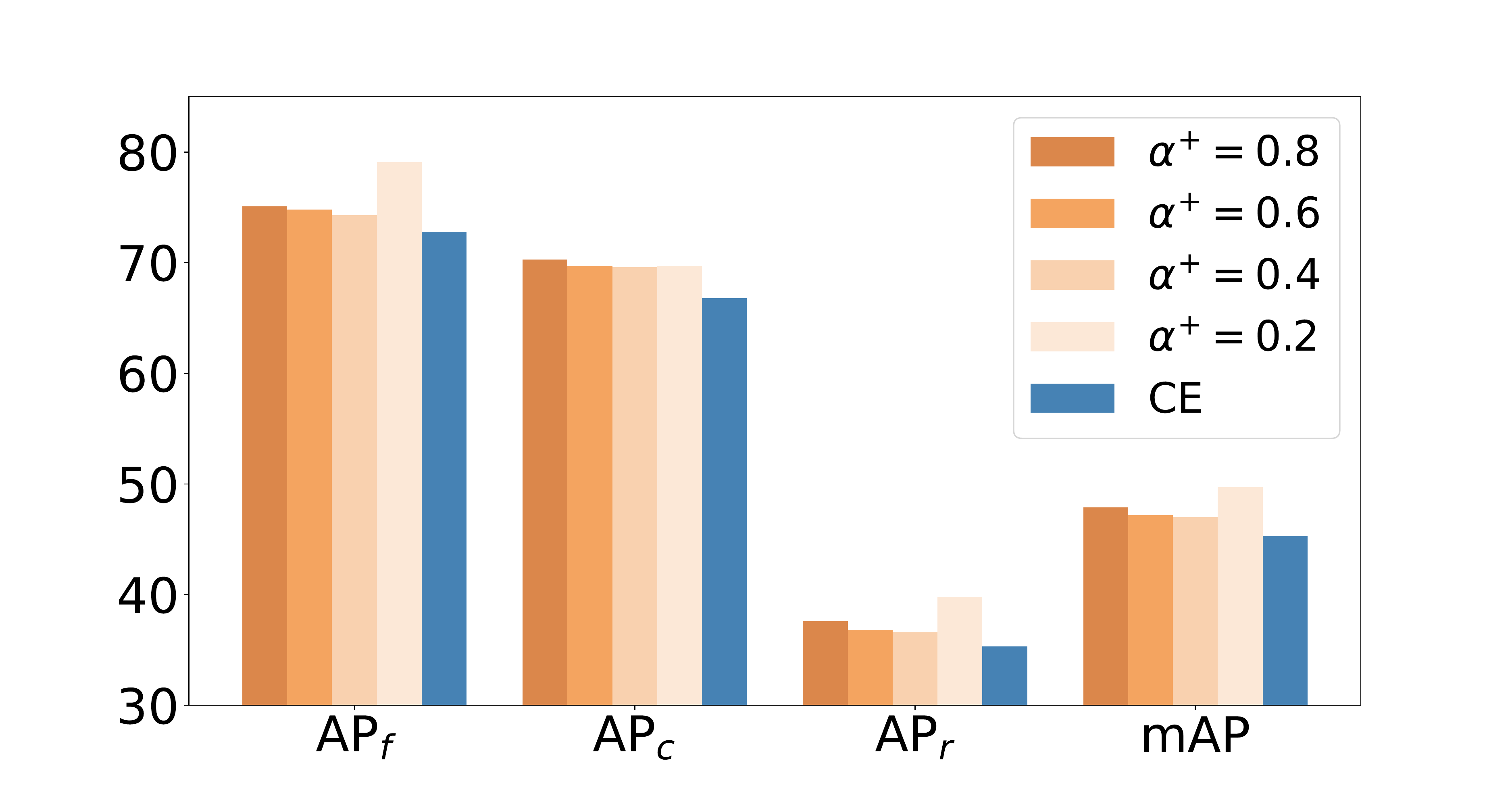}}
    \quad
    \subfigure[]{\includegraphics[width=0.35\textwidth]{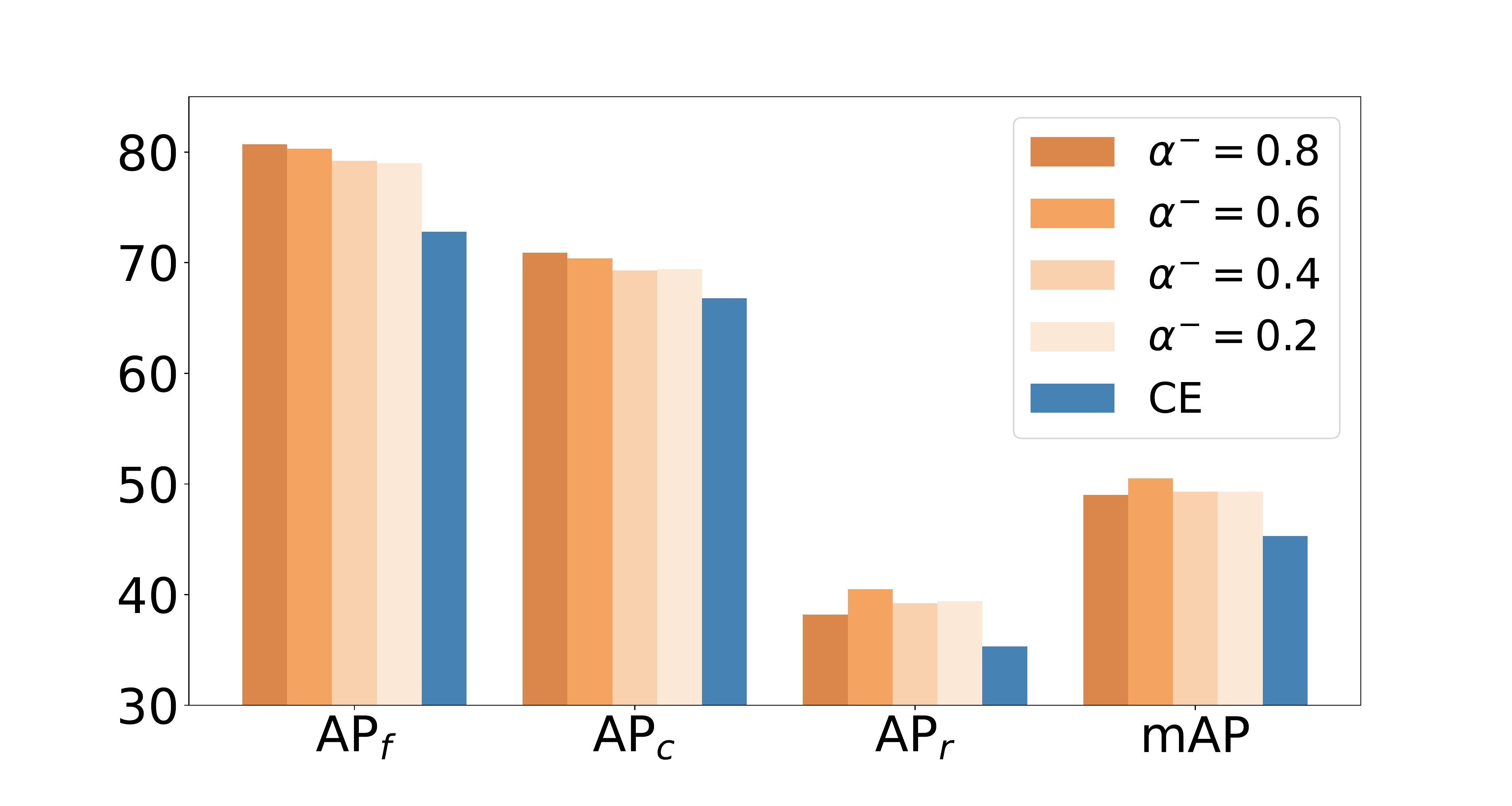}}
    \caption{The effects of $\alpha^{+}$ and $\alpha^{-}$. (a): The individual effects of $\alpha^{+}$ from 0.2 to 0.8. (b): The individual effects of $\alpha^{-}$ from 0.2 to 0.8. CE means the original cross-entorpy loss.}
    \label{fig4}
\end{figure}

\subsection{Performance Analysis}
\textbf{Does our method balance gradients well?} As shown in Fig.\ref{fig3}(a), for Cross-Entropy Loss, the tail classes obtain the gradient imbalance of positive and negative samples, which means that the gradients of positive samples are overwhelmed by the gradients of negative samples. In contrast, our method re-balances the gradients of positive and negative samples in appropriate parameters. \textbf{Does our method balance classifiers well?} Decoupled training methods~\cite{kang2019decoupling,zhou2020bbn} demonstrate that if models are trained with long-tailed datasets, the head classes tend to learn a classifier with larger magnitudes and yields a wider classification boundary in feature space but it hurt data-scarce classes. As shown in Fig.\ref{fig3}(b), the normalized weight norm of the baseline model decreases sharply in tail classes. In contrast, our method provides more balanced classifier weight magnitudes, expands the classification boundary of the tail classes in feature space, and enhances the feature expression of the tail classes.

\begin{table}[t]
\centering
\caption{Effect of combining $\alpha^{+}$ and $\alpha^{-}$ on head and tail classes.}
\label{table:4}
    \centering
    \begin{tabular}{cc|cccccc|c}
    \hline
         \textbf{$\alpha^{+}$} &\textbf{$\alpha^{-}$} & mR$_f$ & AP$_f$ & mR$_c$ & AP$_c$ & mR$_r$ & AP$_r$ & mAP \\ \hline
         -& -&  83.8&	72.8&	80.6&	66.8&	54.3&	35.3&	45.3 \\ \hline
         0.2& 0.6&96.3&	80.3&	94.6&	71.9&	81.1&	42.3&	52.0  \\ 
        0.6& 0.6&96.9&	80.7&	97.2&	72.5&	83.2&	44.2&	52.5 \\ 
        0.8& 0.8& 96.8&	80.1&	98.0&	72.4&	89.9&	43.7& \textbf{53.1} \\  \hline
    \end{tabular}

\end{table}
\begin{table}[t]
\centering
\caption{Performance comparison when Grad-Libra Loss is applied to other detectors.}\label{tab3}
\centering
\begin{tabular}{c|c|cccccc|c}
\hline

Detector & Grad-Libra &  mR$_f$ & AP$_f$ & mR$_c$ & AP$_c$ & mR$_r$ & AP$_r$ & mAP\\  \hline

\multirow{2}{*}{FCOS}~\cite{tian2019fcos} &    &83.1&	70.8&	81.6&	66.2&	55.3&	38.0&	45.9\\
 & \checkmark &92.9&	78.7&	97.2&	72.7&	86.3&	42.2&	51.9 \\\hline
 
\multirow{2}{*}{ATSS}~\cite{zhang2020bridging} &  &79.3&	63.5&	73.7&	60.9&	43.8&	33.5&	42.0\\
 & \checkmark &96.1&	79.1&	96.1&	70.3&	80.4&	41.1&	50.8
 \\\hline
 
\multirow{2}{*}{YOLOF}~\cite{chen2021you} &  &84.9&	64.9&	82.6&	66.1&	42.7&	38.3&	46.5  \\
 & \checkmark &88.4&	68.5&	97.3&	67.8&	85.5&	41.0&	49.1\\
\hline
\end{tabular}
\end{table}

\subsection{Ablation Study}
\textbf{Individual Parameter Contribution.} We study the impact of the individual hyper-parameter in Fig.\ref{fig4}. Both $\alpha^{+}$ and $\alpha^{-}$ can improve the performance of rare, common, and frequent categories. The effect of $\alpha^{+}$ surpasses the baseline by 1.7 to 4.4 mAP. Up-weighting the hard positive samples makes the network focus on the tail classes. The effect of $\alpha^{-}$ exceeds the baseline by 3.7 to 5.2 mAP. $\alpha^{-}$ prevents vast easy negative samples from producing overwhelming loss and dominating the gradients. Notably, the effect of $\alpha^{-}$ is better than $\alpha^{+}$.
 \textbf{Grad-Libra Loss.} By combining both $\alpha^{+}$ and $\alpha^{-}$, the classification performance is further improved, see Table~\ref{table:4}. The performances on rare categories are significantly improved. $\alpha^{+}$ = 0.8 and $\alpha^{-}$ = 0.8 work best overall, achieving a 53.1\% mAP, increasing AP$_f$ by 7.3\%, AP$_c$ by 5.6\% and AP$_r$ by 8.4\% respectively. Our method also significantly increases the recall rate, increasing mR$_f$ by 13\%, mR$_c$ by 17.4\%, and mR$_r$ by 35.6\%, respectively. \textbf{Applied to other detectors.} To demonstrate the generalization ability of Grad-Libra across different detectors, it is applied to FCOS~\cite{tian2019fcos}, ATSS~\cite{zhang2020bridging}, and YOLOF~\cite{chen2021you}, separately. As presented in Table~\ref{tab3}, Grad-Libra also performs well on all those detectors. The overall improvements for FCOS, ATSS, and YOLOF are 6\%, 8.8\%, and 3.5\%, respectively. For rare categories, the mean recall of FCOS, ATSS, and YOLOF increases by 31\%, 36.6\%, and 42.8\%, respectively.

\section{Conclusion}
In this work, we focus on the long-tailed class imbalance problem in cervical cancer detection scenario. We propose a Grad-Libra Loss that leverages the gradients to dynamically calibrate the degree of hardness of each sample for different categories and re-balance the gradients of positive and negative samples. Extensive experiments show that our method obtains better performance compared with other state-of-the-art methods.

\section*{Acknowledgments}
This work was supported in part by the National Natural Science Foundation of China (Grant No. 91959108) and National Natural Science Foundation of China (No. 61973221).

{
\bibliographystyle{splncs04}
\bibliography{paper710}
}

\end{document}